\documentclass{tlp}
\usepackage{multirow}

\usepackage{booktabs}
\usepackage{amsmath}
\usepackage{graphicx}
\usepackage{multirow}
\usepackage{asplisting}
\usepackage{times}
\usepackage{url}
\usepackage{todonotes}
\usepackage{caption}
\begin{document}

\lefttitle{Bruno et. al}

\jnlPage{1}{8}
\jnlDoiYr{2021}
\doival{10.1017/xxxxx}

\title[Improving ASP-based ORS Schedules through Machine Learning Predictions]{Improving ASP-based ORS Schedules through \\ Machine Learning Predictions}

\begin{authgrp}
\author{\gn{Pierangela} \sn{Bruno}}
\affiliation{DeMaCS, University of Calabria, Italy}
\author{\gn{Carmine} \sn{Dodaro}}
\affiliation{DeMaCS, University of Calabria, Italy}
\author{\gn{Giuseppe} \sn{Galat\`a}}
\affiliation{SurgiQ srl, Italy}
\author{\gn{Marco} \sn{Maratea}}
\affiliation{DeMaCS, University of Calabria, Italy}
\author{\gn{Marco} \sn{Mochi}}
\affiliation{SurgiQ srl, Italy}
\end{authgrp}

\history{\sub{xx xx xxxx;} \rev{xx xx xxxx;} \acc{xx xx xxxx}}

\maketitle

\begin{abstract}

The Operating Room Scheduling (ORS) problem deals with the optimization of daily operating room surgery schedules. It is a challenging problem % given it is 
subject to many constraints, like to determine the starting time of different surgeries and allocating the required resources, including the availability of beds in different department units. Recently, solutions to this problem based on Answer Set Programming (ASP) have been delivered. Such solutions are overall satisfying but, when applied to real data, they can currently only verify whether the encoding aligns with the actual data and, at most, suggest alternative schedules that could have been computed. As a consequence, it is not currently possible to generate provisional schedules. Furthermore, the resulting schedules are not always robust.

In this paper, we integrate inductive and deductive techniques for solving these issues. We first employ machine learning algorithms to predict the surgery duration, from historical data, to compute provisional schedules. Then, we consider the confidence of such predictions as an additional input to our problem and update the encoding correspondingly in order to compute more robust schedules. Results on historical data from the ASL1 Liguria in Italy confirm the viability of our integration.
Under consideration in Theory and Practice of Logic Programming (TPLP).
\end{abstract}


\section{Introduction}
The Operating Room Scheduling (ORS) problem consists of optimizing daily surgical schedules in operating rooms. It is a complex and highly constrained problem that requires, among other tasks, determining the starting times of surgeries and allocating the necessary resources~\citep{abedini2016operating,aringhieri_two_2015,DBLP:journals/cor/HamidNWSZ19,DBLP:journals/dss/MeskensDH13}. 
Recently, several solutions based on Answer Set Programming (ASP) for the ORS problem have been proposed~\citep{DBLP:journals/tplp/DodaroGKMP22,DBLP:journals/logcom/DodaroGGMMMS24}, showing promising results in finding feasible and efficient schedules under realistic constraints. However, when applied to real-world data, such as those provided by ASL1 Liguria in Italy~\citep{DBLP:journals/logcom/DodaroGGMMMS24}, these solutions rely on the assumption that surgery durations are known in advance. Specifically, since the scheduling was performed on past surgeries, the actual durations were already available. This allowed researchers to compare the ASP-generated schedules with those historically adopted by the hospital and evaluate potential improvements retrospectively. Nevertheless, in a practical setting, surgery durations are not known beforehand, and this uncertainty poses a critical challenge: ASP systems heavily depend on accurate input values, and imprecise duration estimations can lead to suboptimal scheduling solutions. As a result, the ability to generate provisional schedules under uncertainty remains largely unaddressed.

To overcome this limitation, it is necessary to integrate predictive models capable of estimating surgery durations before the actual scheduling process takes place. Machine learning (ML) techniques offer a promising solution for this purpose, enabling the estimation of surgery durations based on historical patient and surgery data.

The integration of deductive (logic-based) and inductive (ML-based) approaches has emerged as one of the most active areas of research in the AI community in recent years, and ASP is no exception.
Indeed, several efforts have been made in this direction, such as using ML to guide the heuristics of ASP solvers to improve performance~\citep{DBLP:journals/aicom/Balduccini11,DBLP:conf/lpnmr/DodaroIOR22,DBLP:conf/lpnmr/LiuTL22}, applying algorithm selection techniques~\citep{DBLP:journals/tplp/MarateaPR14,DBLP:journals/tplp/HoosLS14}, representing and explaining ML models via ASP~\citep{DBLP:conf/ijcai/EiterGHO23,DBLP:journals/tplp/GiordanoD22}, and developing languages and tools for learning ASP programs~\citep{DBLP:conf/ijcai/YangIL20,DBLP:conf/aaai/TarzariolGSL23,DBLP:conf/ijcai/CunningtonL0R23,DBLP:journals/corr/abs-2005-00904}. In addition, there has been growing interest in neuro-symbolic approaches in real-world applications where ASP is applied in conjunction with ML techniques~\citep{DBLP:journals/tplp/BarbaraGLMQRR23,DBLP:journals/tplp/EiterHOP22,DBLP:conf/lpnmr/BrunoCM22}.

In this paper, we contribute to this line of research by proposing a hybrid approach that integrates ML predictions into an ASP-based solution for the ORS problem. Specifically, our contributions are as follows.
First, we perform an analysis of the available real-world dataset, identifying significant distribution skewness that could negatively affect predictive accuracy. To mitigate this, we apply a dedicated preprocessing phase to improve data quality and reliability.
Then, we systematically evaluate several state-of-the-art ML algorithms for predicting surgery durations, using standard performance metrics such as mean absolute error, root mean squared error, and coefficient of determination. Among the tested models, XGBoost~\citep{chen2016xgboost} achieves the best performance and is selected for further integration.
Subsequently, we introduce the notion of prediction confidence by clustering the predicted durations into four discrete levels, ranging from high confidence to very low confidence, providing an additional layer of information to assess prediction reliability.
Then, we extend the original ASP encoding to incorporate confidence information into the scheduling process, enabling the computation of more robust and reliable surgical schedules.
Finally, we conduct an extensive experimental evaluation. Despite the challenges posed by the inherent distribution skewness in the dataset, and the limited predictive accuracy of the models, the results show that incorporating ML predictions, especially when combined with confidence information, leads to a good improvement in scheduling quality. 
In fact, our approach obtains a better operating room usage and reduces the incidence of overbooking compared to the baseline ASP encoding that relies only on statistical averages.

\section{Problem and data description}
\label{sec:descr}
In this section, we present a high-level description of the ORS problem considered in this paper and the available dataset.

The ORS problem specifications described in the following were defined by ASL1 Liguria, a local health authority in Italy that includes three hospitals: Bordighera, Sanremo, and Imperia. Such hospitals serve a population of around $213,000$ people.
The central element of the ORS problem is the concept of a \textit{registration}. Each registration represents a surgical procedure requested by a patient and is associated with a specific duration, a reservation number, a medical specialty, and a type of hospitalization.
The set of registrations that have not yet been performed constitutes the surgical waiting list.
The overall goal of the ORS problem is to assign as many registrations as possible from a large waiting list to appropriate operating rooms (ORs). Due to resource or specialty constraints, it may not be possible to assign certain surgical specialties to specific ORs. 
Therefore, the objective is to maximize the usage of OR time, which is an extremely valuable resource. Indeed, OR costs are estimated to be in the range of tens of dollars per minute \citep{smith2022ORcosts}, with approximately half representing fixed costs incurred even when the OR is not used \citep{MACARIO2010233}.
Since patients cannot overlap within the same OR and OR overload must be avoided, the first requirement is to ensure that the total duration of surgeries assigned to any given OR does not exceed its available operating time. In the context of the three hospitals managed by ASL1 Liguria, Bordighera has two ORs available from 07:30 A.M. to 01:30 P.M., while Imperia and Sanremo each have five ORs available from 07:30 A.M. to 08:00 P.M.
The ORS problem also includes aspects related to patient prioritization and OR usage. In particular, registrations may correspond to different clinical conditions and, in general, have been placed on the waiting list for varying lengths of time. These factors are jointly abstracted into a unified \textit{priority} metric that guides the scheduling process, where registrations with the highest priority refer to patients already preplanned by the hospital and are therefore subject to a hard constraint, i.e., they must be scheduled, while other registrations are scheduled according to OR capacity and subject to a hierarchical preference.
Furthermore, some ORs are designated for limited elective use due to their partial reservation for emergency procedures or other institutional needs.

More precisely, given a set of surgery registrations (each consisting of a patient ID, a priority level from $p_1$ to $p_4$, the required specialty, and the expected surgery duration) and a Master Surgical Schedule (MSS), which defines the specialty assigned to each OR during each shift of the week, the goal is to assign each registration (i.e., each patient) to an operating room, within a specific shift on a given day of the scheduling period, subject to the following constraints:
\begin{itemize}
\item Each registration is assigned at most once;
\item The total length of the surgeries assigned to a given OR and shift must not exceed the length of the shift;
\item Registrations with priority level $p_1$ must be assigned to an operating room within the scheduling period;
\item Unassigned registrations with priority levels $p_2$, $p_3$, and $p_4$ should be minimized, giving precedence to higher-priority cases;
\item A specific operating room, referred to as \textit{OR A}, can be assigned to at most one patient, as it is reserved for emergencies.
\end{itemize}

As for the datasets, we used data taken from a weekly schedule of surgeries across the three hospitals of ASL1, as well as data from historical other weeks, including a list of available ORs for all hospitals.
We collected and prepared the data for testing by working with four different files, where each file represents a different type of data.
In particular, the first file contains the operating list of the considered week of surgeries, from 04/03/2019 to 10/03/2019, providing information on the required surgery, the operating room, and the specialty originally scheduled.
The second file includes the historical list of surgeries scheduled in 2019, which includes information on the required surgery, the starting and ending time of the surgery, and the date of the surgery.
The third file includes the list of ORs in each hospital and their opening hours.
The fourth file contains the list of patients hospitalized the week before the considered week of the scheduling, along with their admission and discharge times.
%The fifth file contains the list of beds in each specialty at each hospital, which, however, were not used in our case.
%
Overall, the dataset contains 32 features. Each feature represents a different attribute related to the surgical procedure, patient, diagnosis, timing, or logistics.
Each row corresponds to a single surgical intervention. Specifically, Table \ref{tab:dataset_description} shows all the features included in the dataset along with the corresponding description.

\begin{table}[t]
\centering
\caption{Description of the features in the surgical procedures dataset.}
\footnotesize
\begin{tabular}{ll}
\hline\hline
\textbf{Feature} & \textbf{Description}
\midline
\texttt{PROGRESSIVO} & Unique identifier for each surgical record. \\
\texttt{TIPORICOVERO} & Type of hospitalization (e.g., emergency, urgent post-operative). \\
\texttt{SESSO} & Biological sex of the patient (F = Female, M = Male). \\
\texttt{ETA} & Patient age at the time of procedure, expressed in years. \\
\texttt{REPARTO} & Hospital department or unit where the procedure was performed. \\
\texttt{PRES ANESTES} & Presence of anaesthesiologist during the procedure (Yes/No). \\
\texttt{STAMP} & Internal administrative flag. \\
\texttt{CC} & Internal compliance tracking flag. \\
\texttt{CA} & Additional clinical flag. \\
\texttt{ANESTLOC} & Whether localized anesthesia was used. \\
\texttt{DIAGNOSI1} & Primary diagnostic code (ICD format). \\
\texttt{DESCDIAGNOSI1} & Description of the primary diagnosis. \\
\texttt{INGRESSOSALA} & Timestamp of patient entry into the operating room. \\
\texttt{USCITASALA} & Timestamp of patient exit from the operating room. \\
\texttt{REGRICOVERO} & Type of hospital admission (e.g., ordinary, day surgery). \\
\texttt{CHIRURGHI\_1} & Identifier for the surgeon or surgical team. \\
\texttt{ICD1} & Code for the surgical procedure (ICD format). \\
\texttt{DESCICD1} & Description of the surgical procedure. \\
\texttt{BLOCCO} & Surgical block location in the hospital. \\
\texttt{DATANASCITA} & Patient date of birth. \\
\texttt{NOSOLOGICO} & Internal hospital case reference number. \\
\texttt{DATAINTERVENTO} & Date of surgical intervention. \\
\texttt{SALA} & Operating room designation. \\
\texttt{TIPOANESTESIA} & Type of anaesthesia administered. \\
\texttt{INGRESSOBLOCCOOP} & Timestamp of entry into the operative block. \\
\texttt{PREPARAZIONEPAZIENTE} & Start time of preoperative patient preparation. \\
\texttt{INIZIOANESTESIA} & Start time of anesthesia administration. \\
\texttt{INIZIOINTERVENTO} & Start time of the surgical procedure. \\
\texttt{FINEINTERVENTO} & End time of the surgical procedure. \\
\texttt{FINEASSANESTINSALA} & End of anesthesiology assistance in the operating room. \\
\texttt{USCITABLOCCOOP} & Timestamp of patient exit from operative block. \\
\texttt{DURATA} & Total procedure duration in minutes.\\
\hline\hline
\end{tabular}
\label{tab:dataset_description}
\end{table}

\section{Prediction methodology and experimental evaluation}
\label{sec:ml}
This section presents the preprocessing applied to the dataset (Section~\ref{sec:dataprocessing}), the ML algorithms employed to analyze the data (Section~\ref{sec:predictivemodelling}), quantitative model performances (Section~\ref{sec:evaluationmetrics}), and the results of the experiments in terms of confidence (Section~\ref{sec:confidenceestimation}).

\subsection{Data preprocessing and distribution analysis} \label{sec:dataprocessing}
\begin{figure}[t]
    \centering
    \begin{minipage}{0.45\textwidth}
        \includegraphics[width=\linewidth]{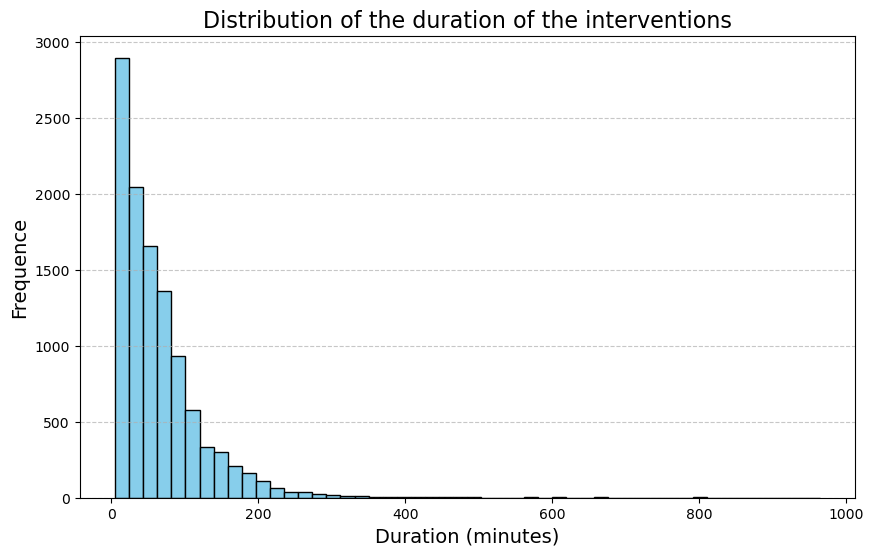}
        \caption{Histogram of intervention durations (before preprocessing).}
        \label{fig:hist_before}
    \end{minipage}
    \begin{minipage}{0.45\textwidth}
        \includegraphics[width=\linewidth]{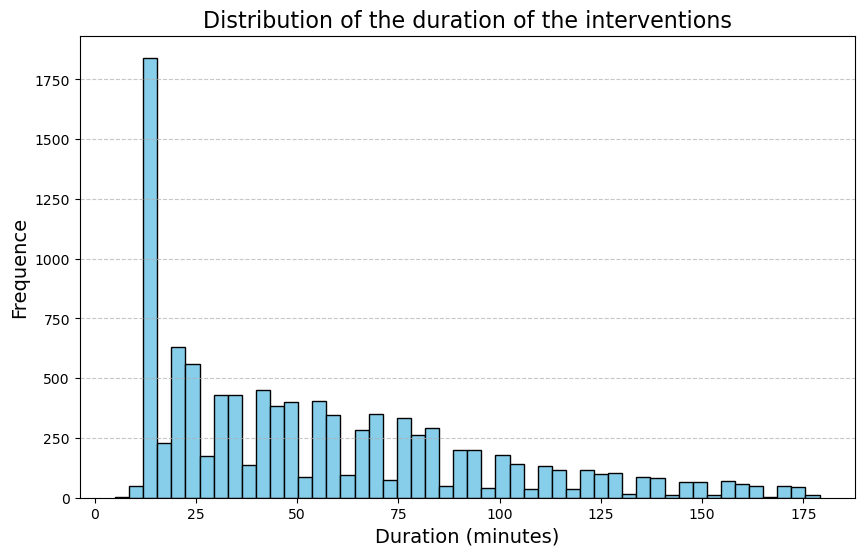}
        \caption{Histogram of intervention durations (after preprocessing).}
        \label{fig:hist_after}
    \end{minipage}
\end{figure}

Data preprocessing is a crucial step in ML, particularly when working with real-world medical datasets that often suffer from skewed distributions, noise, and sparsity. 
In our study, the target variable, \texttt{DURATA}, representing the total operative time in minutes, was derived from the difference between \texttt{USCITASALA} (exit from operating room) and \texttt{INGRESSOSALA} (entry). 
Both fields were parsed as \texttt{datetime} objects, and duration was computed as the difference in minutes. 
Entries with negative or zero durations, likely due to data entry errors, were removed from the dataset to ensure data integrity and modeling accuracy. As the goal of our study is regression, addressing target distribution characteristics is particularly important.
As shown in Figure~\ref{fig:hist_before}, the original distribution of the intervention durations was highly skewed, with a large concentration of cases around 15 minutes and a long tail of less frequent, prolonged procedures. This distribution skewness can significantly affect model performance, leading to biased predictions and poor generalization, particularly for minority classes or rare cases.
To address these issues, we employed the following preprocessing steps:
\begin{enumerate}
    \item Diagnoses that appeared only once within each department (\texttt{REPARTO}) were identified and grouped by K-Means clustering (with up to 3 clusters). This step reduced data sparsity and helped generalize rare cases by associating them with similar groups according to the department.
    \item We filtered out extreme values that lay outside the typical range of durations using the Interquartile Range (IQR) method, that is a statistical measure of dispersion, calculated as the difference between the third quartile (Q3) and the first quartile (Q1) of the data (i.e., IQR = Q3 – Q1) \citep{DekkingEtAl2005}.
    This improved the central tendency and distribution spread, enhancing the learning stability of regression models.
    \item To mitigate issues related to multicollinearity, we removed features that exhibited high pairwise correlation (above a 0.95 threshold). Redundant features can introduce noise, inflate model complexity, and impair generalization. By identifying and eliminating highly correlated variables, we reduced dimensionality and improved model robustness and interpretability. Indeed, by removing highly correlated features, we aimed to reduce redundancy in the dataset. When two or more features are strongly correlated, they tend to capture the same underlying information, making it difficult to isolate their individual contributions to the model. This redundancy complicates interpretation especially when assessing which variables are truly driving predictions, and reduces model stability. As discussed in~\cite{kuhn2013data}, eliminating one of several highly correlated features typically does not impair predictive performance and can lead to a simpler, more interpretable model.
\end{enumerate}
In particular, Figure~\ref{fig:hist_after} illustrates the duration distribution after Step 2 of preprocessing. Compared to the raw dataset, the cleaned data shows a more compact distribution with fewer extreme outliers and a less pronounced right skew.
Furthermore, the preprocessing performed in Step 3 resulted in a reduction of feature dimensionality from 32 to 23, reflecting the elimination of less informative features.
\subsection{Predictive modelling}\label{sec:predictivemodelling}
To predict surgical procedure durations, several ML regressors were implemented and compared. All models were trained and evaluated under identical preprocessing and evaluation pipelines to ensure fairness. 
The dataset was split into training and testing subsets (80\% and 20\%, respectively) using stratified sampling based on procedure duration. 
This strategy ensured that the distribution of durations, particularly for rare or long procedures, was preserved across both sets. 
In this way, we avoided scenarios where specific duration ranges (e.g., particularly long procedures) appeared only in the test set, which could otherwise result in biased or unreliable performance estimations.
The following algorithms were explored:
\begin{itemize}
    \item \textbf{Decision Tree Regressor (DT):} a simple, interpretable model that recursively splits the data based on feature thresholds to minimize prediction error \citep{breiman1984classification}. 
    \item \textbf{Random Forest Regressor (RF):} an ensemble method that constructs multiple decision trees using bootstrap sampling and random feature selection at each split \citep{breiman2001random}. Final predictions are computed as the average of individual tree predictions.
    
    \item \textbf{Gradient Boosting Regressor (GB):} a sequential ensemble technique where each new tree is trained to correct the residuals of the previous ensemble \citep{friedman2001greedy}. It combines weak learners into a strong predictor and includes hyperparameters for controlling learning rate, tree depth, and regularization.

    \item \textbf{Extreme Gradient Boosting (XGBoost):} a highly optimized and regularized gradient boosting framework \citep{chen2016xgboost}, known for its scalability and efficiency. XGBoost introduces system-level optimizations (e.g., parallelization) and algorithmic enhancements like shrinkage and sparsity-aware split finding.
    
    \item \textbf{K-Nearest Neighbors Regressor (KNN):} a non-parametric method that predicts the target as the average of the $k$ nearest neighbors in the feature space \citep{altman1992knn}.
    
    \item \textbf{Support Vector Regressor (SVR):} a margin-based regression technique that aims to find a function within an $\epsilon$-tube from the true outputs, penalizing predictions only when errors exceed $\epsilon$ \citep{drucker1997support}. % It supports both linear and nonlinear kernels.

\end{itemize}

\paragraph{Deep learning (DL) models.}
DL models are generally not well-suited for tabular data, as highlighted by \cite{DBLP:journals/inffus/Shwartz-ZivA22} and by \cite{DBLP:journals/tnn/BorisovLSHPK24}.
Nevertheless, some of the recent DL approaches obtained good performance on tabular data, as the one proposed by \cite{DBLP:conf/aaai/ArikP21}, called TabNet.
Indeed, TabNet is a deep neural architecture tailored for tabular data that uses sequential attention to select features at each decision step, enabling both high performance and interpretability.
Motivated by this promising finding, we tested the performance of TabNet on our dataset and we adapted and evaluated the classical DL models \textbf{Multi-Layer Perceptron (MLP)~\citep{almeida2020multilayer} and 1D Convolutional Neural Network (1D-CNN)}~\citep{DBLP:journals/access/IgeS24}.
Each model was optimized via grid search and the best configuration is reported in Table~\ref{tab:best_params}.
In more detail, MLP was implemented as a feedforward neural network with the Rectified Linear Unit (ReLU), a nonlinear activation function that introduces nonlinearity into the model and helps mitigate the vanishing gradient problem. It outputs the input directly if it is positive and zero otherwise, effectively retaining only the positive part of its argument. We also used mean squared error loss and Adaptive Moment Estimation (Adam) optimizer to make training faster and more stable, especially on noisy or sparse data~\citep{DBLP:journals/corr/KingmaB14}. Our implementation followed a standard configuration with two hidden layers and 64 units per layer.
1D-CNN was adapted to tabular input by reshaping the feature vectors as sequences, applying a single convolutional layer with 32 filters and kernel size 2, followed by a fully connected hidden layer and regression output. 
TabNet was tested using a flexible wrapper that allowed us to tune key architectural parameters such as the size of internal layers, the number of steps in the attention process, and the strength of feature sparsity. 
Unlike standard neural networks, TabNet learns which features to use at each step, making it more efficient for tabular tasks.

\subsection{Evaluation metrics and experiments}\label{sec:evaluationmetrics}
To evaluate the performance of the ML and DL models in predicting surgical procedure durations, we employed three standard regression metrics. The first one is the \textbf{Mean Absolute Error (MAE)} that reflects the average magnitude of prediction errors, and is computed as the average of the absolute differences between the predicted values $\hat{y}_i$ and the true values $y_i$, i.e., $\text{MAE} = \frac{1}{n} \sum_{i=1}^{n} |y_i - \hat{y}_i|$.
The second metric is the \textbf{Root Mean Squared Error (RMSE)}, which penalizes larger errors more than \textbf{MAE} and reflects the model's sensitivity to outliers. RMSE is calculated as the square root of the mean squared error, i.e., $\text{RMSE} = \sqrt{ \frac{1}{n} \sum_{i=1}^{n} (y_i - \hat{y}_i)^2 }$. The last one is the \textbf{Coefficient of Determination (R\textsuperscript{2})}, which measures the proportion of variance in the target variable explained by the model, that is R\textsuperscript{2} $ = 1 - \frac{ \sum_{i=1}^{n} (y_i - \hat{y}_i)^2 }{ \sum_{i=1}^{n} (y_i - \bar{y})^2 }$, where $\bar{y}$ is the mean of the observed values.
%\subsection{Experimental Results}
In addition to model evaluation, we conducted hyperparameter tuning using grid search with cross-validation. The final configuration for each model was selected based on the combination of hyperparameters that achieved the lowest MAE and RMSE averaged across the validation folds during cross-validation, as shown in Table~\ref{tab:best_params} (for details about the hyperparameters see \url{https://github.com/DeMaCS-UNICAL/ML4ORS}).
Moreover, Table~\ref{tab:model_perf} highlights the performance of each ML model in predicting surgical procedure durations. Among all candidates, the XGBoost Regressor achieved the best results, with the lowest MAE (12.36 minutes), lowest RMSE (18.20 minutes), and highest R\textsuperscript{2} score (0.79). RF and GB followed closely behind, also demonstrating strong predictive performance with slightly higher error metrics.
In contrast, simpler models such as DT, KNN, and SVR showed substantially lower performance. The DT and KNN regressors yielded higher MAE values (16.99 and 15.69 minutes, respectively) and lower R\textsuperscript{2} scores (0.55 and 0.64), indicating limited generalization. The SVR model performed comparably to the DT, with the highest MAE (17.22 minutes) and RMSE (26.87 minutes), and an R\textsuperscript{2} of 0.55, suggesting both struggled to capture the underlying structure of the data.

\begin{table}[t!]
\centering
\footnotesize
\caption{Best Hyperparameter Configuration for Each Algorithm.}
\label{tab:best_params}
\begin{tabular}{ll}
\hline\hline
\textbf{Algorithm} & \textbf{Hyperparameter}
\midline
DT & \texttt{max\_depth=50}, \texttt{min\_samples\_split=2}, \texttt{criterion=friedman\_mse}\\
RF & \texttt{n\_estimators=10}, \texttt{max\_depth=None}, \texttt{min\_samples\_split=5}\\
GB & \texttt{n\_estimators=400}, \texttt{learning\_rate=0.01}, \texttt{max\_depth=15}\\
XGBoost & \texttt{n\_estimators=400}, \texttt{learning\_rate=0.1}, \texttt{max\_depth=5} \\
KNN & \texttt{n\_neighbors=5}, \texttt{weights=distance} \\
SVR & \texttt{C=1}, \texttt{epsilon=0.01}, \texttt{kernel=rbf}\\
MLP & \texttt{epochs=100}, \texttt{batch\_size=64} \\
1D-CNN & \texttt{epochs=50}, \texttt{batch\_size=8} \\
TabNet & \texttt{n\_d=8}, \texttt{n\_a=16}, \texttt{n\_steps=3}, \texttt{gamma=2.0},\\
& \texttt{lambda\_sparse=0.0001}, \texttt{lr=0.01} \\
\hline\hline
\end{tabular}
\end{table}
\begin{table}[t]
\centering
\footnotesize
\caption{Model performance using best parameters. Best results are in bold.}
\label{tab:model_perf}
\begin{tabular}{lrrr}
\hline\hline
\textbf{Model} & \textbf{MAE (min)} & \textbf{RMSE (min)} & \textbf{R\textsuperscript{2}}
\midline
DT           & 16.99 & 26.74 & 0.55 \\
RF         & 12.93 & 19.45 & 0.76 \\
GB        & 13.32 & 20.42 & 0.74 \\
XGBoost                  & \textbf{12.36} & \textbf{18.20} & \textbf{0.79} \\
KNN      & 15.69 & 24.03 & 0.64 \\
SVR & 17.22 & 26.87 & 0.55 \\
MLP & 15.73 & 23.28 & 0.66 \\
1D-CNN & 15.24 & 22.57 & 0.68 \\
TabNet & 14.45 & 22.38 & 0.69 \\
\hline\hline
\end{tabular}
\end{table}

These discrepancies in performance can be partially explained by the distributional characteristics of the target variable. As shown in Figure~\ref{fig:hist_after}, the duration data remains skewed even after preprocessing. 
Most surgeries are short-clustered around 15-20 minutes, but a long tail of high-duration outliers exists. This distribution skewness makes the prediction task inherently challenging, particularly for models like SVR and KNN, which are more sensitive to variance and do not adapt well to skewed or noisy distributions. In contrast, ensemble methods like XGBoost and GB are well-suited to handle skewed data, thanks to their iterative learning approach and ability to capture complex, nonlinear relationships.
It is important to observe that the overall error is still non-negligible (as it is around 12 minutes on average).
Nevertheless, this level of accuracy, combined with the confidence estimation framework (where over 65\% of predictions were classified as High or Moderate confidence) makes the models suitable for potential deployment in time-sensitive hospital operations and to be employed to improve the robustness of the schedules, as we will show in Section~\ref{sec:exp}. 

Concerning DL approaches, it is possible to observe that they do not outperform XGBoost in this task. However, TabNet demonstrates improved performance compared to other DL models, with an MAE of 14.45 minutes, RMSE of 22.38 minutes, and R\textsuperscript{2} of 0.69, which is better than both the 1D-CNN (MAE: 15.24, RMSE: 22.57, R\textsuperscript{2}: 0.68) and MLP (MAE: 15.73, RMSE: 23.28, R\textsuperscript{2}: 0.66).
Nevertheless, despite its competitive performance, TabNet still remains slightly behind XGBoost (MAE: 12.36, RMSE: 18.20, R\textsuperscript{2}: 0.79).

\begin{figure}[t]
    \centering
    \includegraphics[width=0.6\linewidth]{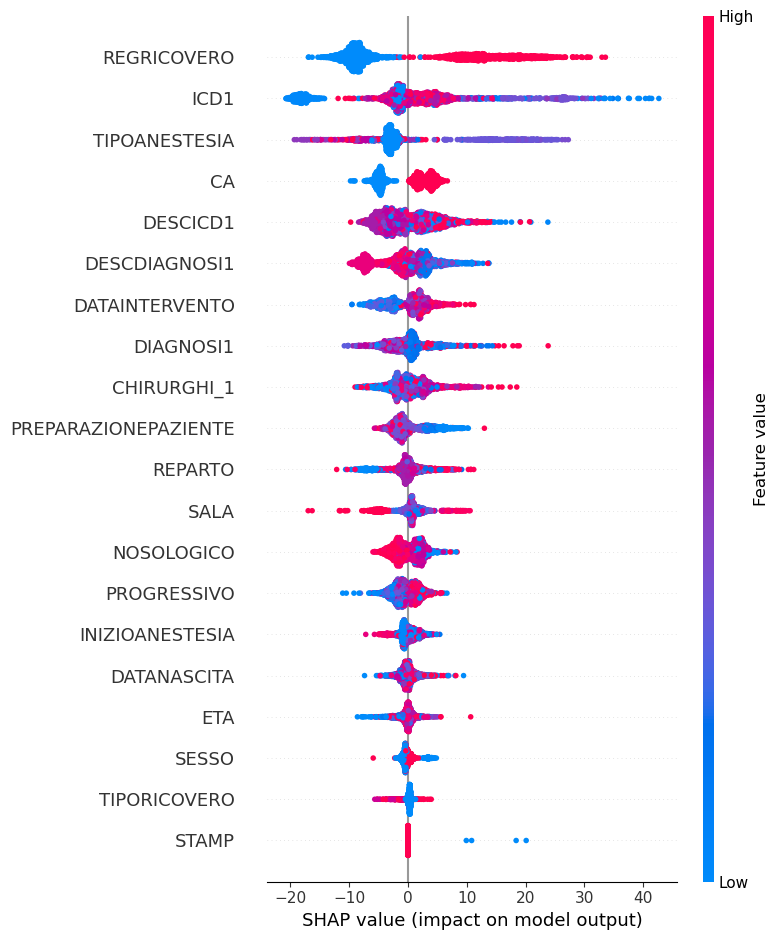}
    \caption{SHAP summary plot for the best-performing regression model (XGBoost).}
    \label{fig:shap_summary}
\end{figure}
Moreover, to interpret the contribution of each feature to the model's predictions, we employed SHAP (SHapley Additive exPlanations) that assigns each feature an importance value for a particular prediction \citep{lundberg2017unified}. 
Figure~\ref{fig:shap_summary} shows the SHAP summary plot for the best-performing model (XGBoost), where features are ranked by their overall contribution to the predictions. The horizontal spread reflects the variability in feature impact across all samples, while the color gradient indicates the feature value (from low in black to high in red).
As an example, \texttt{REGRICOVERO}, representing the type of hospital admission, shows a wide range of SHAP values, highlighting its strong and variable influence on predicted operative duration. Higher values of this feature (in red) tend to increase the predicted duration, whereas lower values (in blue) are associated with shorter procedures. This suggests the model has learned to associate certain admission types (e.g., complex or urgent cases) with longer surgery times.

\subsection{Confidence estimation}\label{sec:confidenceestimation}
\label{subsec:conf}
%To enhance interpretability, i.e., to better understand and communicate the reliability of the model's predictions, we computed the \textbf{Absolute Percentage Error (APE)} for each prediction. 
To assess the precision of individual predictions, we computed the \textbf{Absolute Percentage Error (APE)} for each instance.
This metric quantifies the relative difference between the predicted value $\hat{y}$ and the true target value 
$y$:

\[
\text{APE} = \frac{|\hat{y} - y|}{y} \cdot 100.
\]
In particular, the absolute value ensures that the error is always positive, and the percentage form allows for intuitive interpretation across varying scales.

Based on the APE, we categorized each prediction into one of four confidence levels, namely \textbf{High Confidence} when APE is less than 10\%, \textbf{Moderate Confidence} when APE is between 10\% and 25\%, \textbf{Low Confidence} when APE is between 25\% and 50\%, and \textbf{Very Low Confidence} when APE is greater than or equal to 50\%.

\section{Improving ASP-based ORS solution through ML predictions}
In this section, we first review the ASP encoding of the ORS problem evaluated on real data employed by \cite{DBLP:journals/logcom/DodaroGGMMMS24} (Section~\ref{sec:ors}).
Then, we present the changes needed to take into account ML predictions (Section~\ref{sec:integrationmlasp}).
Finally, we show the results of an experimental analysis that demonstrate the improvements of our approach (Section~\ref{sec:exp}).

In the following sections, we assume the reader is familiar with logic programming conventions, and with ASP syntax and semantics~\citep{DBLP:journals/cacm/BrewkaET11,CalimeriFGIKKLM20}.

\subsection{ASP-based ORS solution}
\label{sec:ors}
In this section, we briefly describe an existing ASP-based encoding for the ORS problem. We begin by describing the problem's input and expected output and then show the core rules that model the scheduling constraints and objectives within the ASP framework. This encoding serves as the baseline for the extensions introduced in the following sections.

\paragraph{\bf Data model.}
The input data is specified by means of the following constants and atoms.
Instances of \lstinline[language=asp]{registration(ID,P,SP,DUR)} represent the registration of the patient identified by an ID (\lstinline[language=asp]{ID}) with priority level (\lstinline[language=asp]{P}), the requested specialty (\lstinline[language=asp]{SP}), and the expected duration of the surgery (\lstinline[language=asp]{DUR}).
Instances of \lstinline[language=asp]{mss(OR,SP,SHIFT,DAY)} represent which specialty (\lstinline[language=asp]{SP}) is assigned to an OR (\lstinline[language=asp]{OR}) in a shift (\lstinline[language=asp]{SHIFT}) on a day (\lstinline[language=asp]{DAY}).
Instances of \lstinline[language=asp]{shift(SHIFT,DURATION)} indicate that the total duration of all surgeries scheduled in shift \lstinline[language=asp]{SHIFT} must not exceed \lstinline[language=asp]{DURATION} (expressed in time slots).
The output is represented by atoms of the form \lstinline[language=asp]{x(ID,P,OR,DAY,SHIFT)},
whose intuitive meaning is that the patient identified by an ID (\lstinline[language=asp]{ID}) having a priority (\lstinline[language=asp]{P}) is assigned to the OR (\lstinline[language=asp]{OR}) in the shift (\lstinline[language=asp]{SHIFT}) on the day (\lstinline[language=asp]{DAY}).

\begin{asp}[float=tp, caption={ASP encoding for the ORS problem.}, label={fig:encoding}]
{x(ID,P,OR,DAY,SHIFT)} :- registration(ID,P,SP,DUR), mss(OR,SP,SHIFT,DAY).
:- #count{OR,DAY,SHIFT: x(ID,P,OR,DAY,SHIFT)} > 1, registration(ID,_,_,_).
:- mss(OR,_,SHIFT,DAY), #sum{DUR, ID: x(ID, _, OR, DAY, SHIFT), registration(ID,_,_,DUR)} > DURATION, shift(SHIFT, DURATION).
:- not x(ID,_,_,_,_), registration(ID,1,_,_).
:~ not x(ID,_,_,_,_), registration(ID,P,_,_). [1@7-P, ID]
:- #count{ID: x(ID,_,_,"OR A",_)} > 1.
\end{asp}
\paragraph{\bf Encoding.}
The related encoding is shown in Listing~\ref{fig:encoding}.
The choice rule assigns an OR, a day, a shift to each registration.
The second rule ensures that each registration is assigned at most once.
The third rule ensures that the total duration of the assigned registration does not exceed the length of the shift.
The fourth rule ensures that every registration with priority $p_1$ is assigned to some OR on a day, shift.
The weak constraint then minimizes the number of unassigned registrations with priority $p_2$, $p_3$, and  $p_4$.
Finally, the last constraint is added to ensure that the specific OR \lstinline[language=asp]{"OR A"} is assigned to at most one patient, as it was reserved for emergencies and used in this limited way in the original data.

\begin{asp}[float=tp, caption={Additional rules to take into account confidence.}, label={fig:add_encoding}]
sumConfidence(DAY, OR, SHIFT, N) :- mss(OR, _, SHIFT, DAY), #sum{L, ID: confidence(ID, L), x(ID, _, OR, DAY, SHIFT)} = N.
maxConfidenceDay(N) :- #max{L: sumConfidence(_, _, _, L)} = N.
minConfidenceDay(N) :- #min{L: sumConfidence(_, _, _, L)} = N.
:~ maxConfidenceDay(N). [N@2]
:~ maxConfidenceDay(MAX), minConfidenceDay(MIN), MAX-MIN > 0. [MAX-MIN@1]
\end{asp}
\subsection{Integration of the ML predictions in ASP}\label{sec:integrationmlasp}
As discussed before, the ORS solutions described by~\cite{DBLP:journals/logcom/DodaroGGMMMS24} can only verify whether the encoding aligns with the actual data and, at most, suggest alternative schedules that could have been computed. Thus, it is not possible to generate provisional schedules.
Furthermore, the resulting schedules are not always robust.
In this section, we describe how this encoding has been extended to incorporate predictive information obtained through ML techniques. The core idea is to exploit the confidence scores produced by the predictive models (specifically, the XGBoost algorithm, which achieved the best results in our evaluation, as shown in Section~\ref{sec:evaluationmetrics}) in order to guide the scheduling decisions made by the ASP program.

To this end, we first update the parameter \lstinline[language=asp]{DUR} of the registrations. Indeed, now the duration is predicted by the ML algorithm.
Then, we also introduce atoms of the form \lstinline[language=asp]{confidence(ID,L)} that collect the confidence associated to each registration, where \lstinline[language=asp]{L} is 1 if the confidence is High, 2 if the confidence is Moderate, 3 if the confidence is Low, and 4 if the confidence is Very Low.

Finally, we handle these two changes by adding a new set of rules, reported in Listing~\ref{fig:add_encoding}.
The key idea is to prefer balanced and high-confidence scheduling decisions.
Indeed, the first rule derives the sum of all the confidences assigned to a particular day.
The subsequent two rules derive the values corresponding to the maximum and the minimum sum of confidences in each day, respectively.
Then, the first weak constraint penalizes solutions where the maximum sum of confidence scores assigned to a day and OR is high, whereas the second one promotes balance across ORs and days by penalizing the difference between the maximum and minimum confidence sums.
In this way, we aim at distributing equally among the ORs and days the patients with high confidence value.

\subsection{Experiments}\label{sec:exp}
This section presents the experimental evaluation conducted to assess the effectiveness of the proposed neuro-symbolic approach.

First of all, it is important to observe that, in the context of ORS, it is not possible to determine the exact duration of a surgical procedure in advance. Typically, the parameter \lstinline[language=asp]{DUR} of the atoms \lstinline[language=asp]{registration(ID,P,SP,DUR)} is assigned by hospital staff based on prior experience. In most cases, it corresponds either to the average duration of surgeries of the same type or to the average duration of surgeries performed within the same department.
Therefore, for the purposes of the following experimental analysis, we compare these two traditional approaches (average by procedure type and average by department) against machine learning-based strategies described in the previous sections. Specifically, we replace the \lstinline[language=asp]{DUR} value in the ASP encoding with the respective predicted duration, compute the surgical schedule using ASP, and then evaluate the quality of the schedules based on the actual durations of the procedures. The real durations, available only after the scheduling has been executed, are crucial for accurately estimating how effectively the operating rooms were used on each day.

We evaluate all the approaches on real-world data provided by ASL1 Liguria and already used by~\cite{DBLP:journals/logcom/DodaroGGMMMS24}. The evaluation focuses on key metrics such as the OR percentage occupancy (mean, standard deviation, minimum, maximum), and the number of times in which an OR has been underbooked (an OR is used less than 80\% of the time) or overbooked (an OR is used for more than 100\% of the time).
It is important to note that, with respect to these parameters, ASP solutions (even within the same solver) may exhibit unintuitive behaviour. For instance, a solution that is preferable in terms of optimality (e.g., lower cost) may still yield worse overbooking and underbooking values, since these aspects are evaluated only in a post-processing phase. To mitigate this issue, in our encoding, the weak constraints related to confidence are assigned a lower priority level than the original ones. This ensures that performance differences can be attributed primarily to the introduction of the new constraints.

The comparison has been carried out on an Apple M1 CPU machine with 8 GB of physical RAM and a time limit of 60 seconds per run. As ASP system, we used {\sc clingo} v. 5.6.2, configured with parameters \textit{-{}-restart-on-model} and \textit{-{}-parallel-mode=6}. These parameters have been found to be effective in a preliminary analysis we performed with several options. 

The results are presented in Table~\ref{tab:results}. The column \textbf{VBA} refers to the virtually best approach, in which the durations are set to their actual values, known only after the surgeries have been performed. This serves as a reference for the optimal performance achievable by predictive methods.
The column \textbf{Conf.} refers to the method in which the parameter \lstinline[language=asp]{DUR} is set using the durations predicted by the machine learning model XGBoost, with the ASP encoding also taking into account the associated confidence information. The column \textbf{Pred.} corresponds to the method where \lstinline[language=asp]{DUR} is set using the XGBoost predictions without considering the confidence scores. The columns \textbf{Dep.} and \textbf{Surg.} represent the methods where \lstinline[language=asp]{DUR} is assigned based on the average duration per department and per surgical procedure, respectively.

\begin{table}[t]
\scriptsize
\centering
\caption{
Comparison of the different methods. 
}
\label{tab:results}
\setlength{\tabcolsep}{1pt}
\begin{tabular}{llllllllllllllllll}
\hline\hline
& \multicolumn{5}{c}{Bordighera} & \phantom{aaaa}& \multicolumn{5}{c}{Imperia} &\phantom{aaaa}& \multicolumn{5}{c}{Sanremo}\\
\cmidrule{2-6} \cmidrule{8-12} \cmidrule{14-18}
& \textbf{VBA} & \textbf{Conf.} & \textbf{Pred.} & \textbf{Dep.} & \textbf{Surg.} && \textbf{VBA} & \textbf{Conf.} & \textbf{Pred.} & \textbf{Dep.} & \textbf{Surg.} && \textbf{VBA} & \textbf{Conf.} & \textbf{Pred.} & \textbf{Dep.} & \textbf{Surg.}
\midline
\% OR occ. (mean) & 99 & 96 & 95 & 101 & 88 && 99 & 100 & 100 & 101 & 102 && 100 & 101 & 101 & 103 & 103\\
OR occ. (std) & 0.87 & 11.32 & 12.64 &  11.25 &  10.35 && 0.68 & 10.32 & 11.16 & 13.14 & 10.21 && 0.51 & 10.21 & 7.8 & 13.24 & 8.65\\
\% OR occ. (max) & 99 & 124 & 120 & 132 &  107 && 99 & 126 & 152 & 136 & 132 && 100 & 118 & 114 & 120 & 119\\
\% OR occ. (min) & 97 & 79 &  67 &  88 &  69 &&  97 & 78 & 80 & 60 & 81 && 99 & 86 & 85 & 74 & 94\\
Overbooking & 0 & 9 & 9 & 11 & 10 && 0 & 40 & 40 & 42 & 44 && 0 & 5 & 9 & 8 & 8\\
Underbooking & 0 & 1 & 2 & 0 & 6 && 0 & 1 & 1 & 2 & 0 && 0 & 0 & 0 & 1 & 0\\
\hline\hline
\end{tabular}
\end{table}

For the evaluation of the scheduling results, it is important to note that the ideal situation is to maximize the usage of ORS without exceeding their available capacity. Thus, an OR occupancy rate close to, but not exceeding, 100\% is preferred. 
Schedules that lead to overbooking (i.e., planned durations exceeding the available OR time) are undesirable, as they can cause operational disruptions and delays. Similarly, underbooking (i.e., leaving significant unused OR time, i.e. below 80\%) should be minimized, although it is generally less critical than overbooking. Therefore, among the generated schedules, the most desirable are those that maintain high, balanced occupancy rates without causing overbooking and with minimal underbooking.

By analyzing the results, we observe that the method \textbf{Conf.} (XGBoost predictions with confidence information) achieves the best overall performance, followed by \textbf{Pred.} (XGBoost predictions without confidence). In Bordighera, the mean OR occupancy with \textbf{Conf.} is 96\%, which is very close to the ideal 100\% threshold, outperforming \textbf{Dep.} (101\%) and \textbf{Surg.} (88\%). Additionally, the standard deviation is reasonably low, indicating consistent scheduling performance across different days.
The number of overbooked cases is slightly lower or comparable to the other methods, and underbooking is almost negligible.
Also \textbf{Pred.} obtains a good performance being quite close to \textbf{Conf.}.
In Imperia, all methods reach a mean occupancy around 100\%, but \textbf{Conf.} maintains better control over the standard deviation compared to \textbf{Dep.} and \textbf{Surg.}, and slightly fewer overbooked cases are observed.
Similarly, in Sanremo, the \textbf{Conf.} and \textbf{Pred.} methods result in a mean occupancy closer to 100\% (101\%), whereas methods like \textbf{Dep.}, and \textbf{Surg.} tend to overbook, with mean occupancies exceeding 103\%. Furthermore, \textbf{Conf.} achieves the lowest number of overbooking (5) compared to all the other methods, and avoids underbooking altogether.

Overall, integrating confidence information into the scheduling process improves the average OR usage and also helps limiting extreme cases of overbooking and underbooking, leading to more balanced and operationally feasible schedules.

\section{Related work}
The integration of symbolic reasoning and machine learning has garnered significant attention in recent years, leading to the emergence of neuro-symbolic AI. This paradigm aims to combine the learning capabilities of neural networks with the interpretability and formal reasoning of symbolic approaches \citep{DBLP:journals/expert/ShethR24}. 
In the context of ASP, neuro-symbolic methods have been explored to enhance various applications, including knowledge representation and visual question answering (VQA).
In particular, \cite{DBLP:journals/tplp/BarbaraGLMQRR23} present a combination of deep learning techniques with ASP, and allows for identifying possible anomalies and errors in the final product of an Italian Company operating in electrical control panel production, which provided real data. 
\cite{DBLP:conf/lpnmr/BrunoCM22} define a framework to represent and solve explicit knowledge via ASP, taking advantage of it for driving decisions taken by neural networks and refining the output for providing explanations and interpretations. The framework has been tested on semantic segmentation tasks over two datasets of biomedical images. 
As for VQA, \cite{DBLP:journals/tplp/EiterHOP22} introduce a neuro-symbolic pipeline for the analysis of CLEVR, which is a well-known dataset that consists of pictures showing scenes with objects and questions related to them. They employ confidence thresholds into the logic programs to be solved by an ASP solver, with the goal of making robust VQA systems. In the same VQA domain, \cite{DBLP:conf/padl/BasuSG20} and \cite{DBLP:journals/firai/RileyS19} also combine ASP and ML.

Concerning the ORS problem, \cite{Gr2018ApplicationOO} provide a comprehensive overview of various approaches, discussing both different solution methods and alternative problem formulations.
Among the works proposing and evaluating solutions on real-world data, \cite{aringhieri_two_2015} address the scheduling of surgical interventions over a one-week planning horizon, considering multiple departments sharing a fixed number of ORs and post-operative beds. The authors propose a two-phase approach aimed at minimizing patient waiting times and maximizing the usage of hospital resources. Their method first generates a feasible assignment and then optimizes the scheduling plan.
Similarly, \cite{landa_hybrid_2016} tackle the ORS problem by decomposing it into two interconnected sub-problems: assigning patients to specific dates within a given planning horizon and determining their allocation and sequencing within the ORs. To solve this, a hybrid two-phase optimization algorithm is introduced, combining neighborhood search techniques with Monte Carlo simulation to efficiently explore the solution space.
Another relevant contribution is provided by \cite{DBLP:journals/cor/HamidNWSZ19}, who incorporate Decision-Making Styles (DMS) of surgical teams to better handle constraints related to material and resource availability, patient priorities, and the competencies of surgical staff. They develop a multi-objective mathematical model and design two metaheuristic algorithms to find Pareto-optimal solutions, which have been validated using data collected from a hospital in Iran.
In a different direction, \cite{zhang_stochastic_2017} address the scheduling of both elective and non-elective patients by introducing a time-dependent policy that prioritizes patients dynamically based on urgency levels and waiting times. The problem is formulated as a stochastic shortest-path Markov Decision Process (MDP) with blind alleys and is solved using an asynchronous value iteration method. Experimental results on synthetic data show that the time-dependent policy significantly reduces patient waiting times compared to classical MDP models, without leading to excessive increases in OR usage.
The ORS problem has also been successfully solved by ASP solutions \citep{DBLP:journals/ia/DodaroGMP19} also in the presence of additional resources, such as beds \citep{DBLP:journals/tplp/DodaroGKMP22} and care units \citep{DBLP:conf/aiia/GalataMMMP21}, demonstrating its flexibility in modeling complex scheduling scenarios with multiple resource constraints. In this work, however, we focus on a different aspect of the problem: we build upon the real-world data presented by \citep{DBLP:journals/logcom/DodaroGGMMMS24}, concentrating specifically on the prediction and integration of surgical procedure durations.
Additional resources, such as post-operative beds or care units, were not explicitly considered in our model, as they do not directly impact the duration of surgical interventions, i.e., the key parameter that our ML models were designed to predict and integrate into the scheduling process.

% Overall, the existing literature suggests the importance of integrating predictive, optimization, and resource-management strategies to address the complexity of the ORS problem. 
Our work is based on a neuro-symbolic approach that leverages ML predictions to enhance the robustness and efficiency of ASP-based scheduling.
To the best of our knowledge, this is the first paper in this direction for the ORS problem.

\section{Conclusion}
\label{sec:conc}
In this paper, we have proposed a neuro-symbolic approach to the ORS problem, combining ML techniques with ASP. Starting from an existing ASP encoding, we extended the model to incorporate the confidence scores produced by an ML predictor, specifically, an XGBoost model trained to estimate surgical intervention durations. By integrating this predictive information into the scheduling process, we aimed to enhance both the robustness and the practical efficiency of the resulting schedules.
Experimental evaluations conducted on real-world data from ASL1 Liguria demonstrated that our approach achieves better operating room usage compared to traditional methods based on historical averages. In particular, leveraging confidence information allowed the ASP solver to generate schedules that are closer to the ideal occupancy threshold reducing overbooking or underbooking issues w.r.t. existing approaches. These results highlight the potential of neuro-symbolic techniques in improving scheduling performance in complex, resource-constrained environments.
As future work, one might take into account additional predictive elements, e.g. patient-specific surgical risks, or apply our approach to other problems. 
Moreover, we also observe that the confidence levels derived from APE thresholds cannot provide uncertainty estimations at inference time, as they are computed a posteriori and require access to the ground truth values. In this paper, our objective was not to quantify model uncertainty, but rather to demonstrate how prediction quality metrics can be integrated into the scheduling optimization. Furthermore, since APE relies on ground truth values, the absence of such data may be addressed by leveraging external estimations of expected durations based on standardized clinical averages per procedure type as reported in the medical literature. Nevertheless, exploring uncertainty quantification techniques capable of providing confidence estimations directly at inference time represents a promising direction for future work.
In particular, since XGBoost emerged as the best-performing approach in our evaluation, it would be interesting to investigate whether using confidence measures specifically tailored to XGBoost could further improve performance. In our preliminary analysis, the version of XGBoost combined with our confidence slightly outperformed the variants using dedicated confidence measures. However, various parameter combinations and configurations remain to be explored. While such an approach might yield better performance, it would come at the cost of reduced generality compared to the method proposed in this paper. Nonetheless, this direction represents a promising avenue for future research.
All material for reproducibility is available at \url{https://github.com/DeMaCS-UNICAL/ML4ORS}.

\section*{Acknowledgments}
Carmine Dodaro and Marco Maratea were supported by the European Union - NextGenerationEU and by Italian Ministry of Research (MUR) under PNRR project FAIR ``Future AI Research'', CUP H23C22000860006 and by the European Union - NextGenerationEU and by the Ministry of University and Research (MUR), National Recovery and Resilience Plan (NRRP), Mission 4, Component 2, Investment 1.5, project ``RAISE - Robotics and AI for Socio-economic Empowerment'' (ECS00000035) under the project ``Gestione e Ottimizzazione di Risorse Ospedaliere attraverso Analisi Dati, Logic Programming e Digital Twin (GOLD)'', CUP H53C24000400006. Carmine Dodaro and Pierangela Bruno were supported by the European Union - NextGenerationEU and by Italian Ministry of Research (MUR) under PNRR project Tech4You ``Technologies for climate change adaptation and quality of life improvement'', CUP H23C22000370006.
The research of Marco Mochi and Giuseppe Galat\`a is partially funded by the ``POR FESR Liguria 2014-2020".

\section*{Conflicts of interest}
The authors declare none.

\bibliographystyle{tlplike}
\bibliography{bibtex}
\end{document}